\documentclass[11pt,a4paper]{article}
\usepackage[hyperref]{naaclhlt2018}
\usepackage{times}
\usepackage{latexsym}
\usepackage{svg}
\usepackage{url}

\aclfinalcopy 

\title{SeerNet at SemEval-2018 Task 1: Domain Adaptation for Affect in Tweets}

\author{
    Venkatesh Duppada, Royal Jain and Sushant Hiray \\
    Seernet Technologies, LLC \\
    \{\tt{venkatesh.duppada, royal.jain, sushant.hiray\}@seernet.io}
}

\date{}

\begin{document}
\maketitle
\begin{abstract}
  The paper describes the best performing system for the SemEval-2018 Affect in Tweets (English) sub-tasks. The system focuses on the ordinal classification and regression sub-tasks for valence and emotion. For ordinal classification valence is classified into 7 different classes ranging from -3 to 3 whereas emotion is classified into 4 different classes 0 to 3 separately for each emotion namely anger, fear, joy and sadness. The regression sub-tasks estimate the intensity of valence and each emotion. The system performs domain adaptation of 4 different models and creates an ensemble to give the final prediction. The proposed system achieved 1\textsuperscript{st} position out of 75 teams which participated in the fore-mentioned sub-tasks. We outperform the baseline model by margins ranging from 49.2\% to 76.4 \%, thus, pushing the state-of-the-art significantly. 
\end{abstract}

\section{Introduction} \label{introduction}
Twitter is one of the most popular micro-blogging platforms that has attracted over 300M daily users\footnote{https://www.statista.com/statistics/282087/number-of-monthly-active-twitter-users/} with over 500M \footnote{http://www.internetlivestats.com/twitter-statistics/} tweets sent every day. Tweet data has attracted NLP researchers because of the ease of access to large data-source of people expressing themselves online. Tweets are micro-texts comprising of emoticons, hashtags as well as location data, making them feature rich for performing various kinds of analysis. Tweets provide an interesting challenge as users tend to write grammatically incorrect and use informal and slang words. 

In domain of natural language processing, emotion recognition is the task of associating words, phrases or documents with  emotions from predefined using psychological models. The classification of emotions has mainly been researched from two fundamental viewpoints. \cite{ekman1992argument} and \cite{plutchik2001nature} proposed that emotions are discrete with each emotion being a distinct entity. On the contrary, \cite{mehrabian1980basic} and \cite{russell1980circumplex} propose that emotions can be categorized into dimensional groupings.

Affect in Tweets \cite{SemEval2018Task1} - shared task in SemEval-2018 focuses on extracting affect from tweets confirming to both variants of the emotion models, extracting valence (dimensional) and emotion (discrete). Previous version of the task \cite{mohammad2017wassa} focused on estimating the emotion intensity in tweets. We participated in 4 sub-tasks of Affect in Tweets, all dealing with English tweets. The sub-tasks were: EI-oc: Ordinal classification of emotion intensity of 4 different emotions (anger, joy, sadness, fear), EI-reg: to determine the intensity of emotions (anger, joy, sadness, fear) into a real-valued scale of 0-1, V-oc: Ordinal classification of valence into one of 7 ordinal classes [-3, 3], V-reg: determine the intensity of valence on the scale of 0-1.

Prior work in extracting Valence, Arousal, Dominance (VAD) from text primarily relied on using and extending lexicons \cite{bestgen2012checking} \cite{turney2011literal}. Recent advancements in deep learning have been applied in detecting sentiments from tweets \cite{tang2014learning}, \cite{liu2012emoticon}, \cite{mohammad2013nrc}. 

In this work, we use various state-of-the-art machine learning models and perform domain adaptation \cite{pan2010survey} from their source task to the target task. We use multi-view ensemble learning technique \cite{kumar2016multi} to produce the optimal feature-set partitioning for the classifier. Finally, results from multiple such classifiers are stacked together to create an ensemble \cite{polikar2012ensemble}.

In  this  paper,  we  describe  our  approach and experiments to solve this problem. The rest  of  the  paper  is  laid  out  as  follows:  Section \ref{description} describes the system architecture, Section 
\ref{result} reports  results  and  inference  from  different  experiments. Finally we conclude in Section \ref{conclusion} along with a discussion about future work.

\section{System Description}\label{description}

\subsection{Pipeline} \label{pipeline}
\begin{figure*}[!htbp]
	\center
	\includegraphics[width=1.0\textwidth]{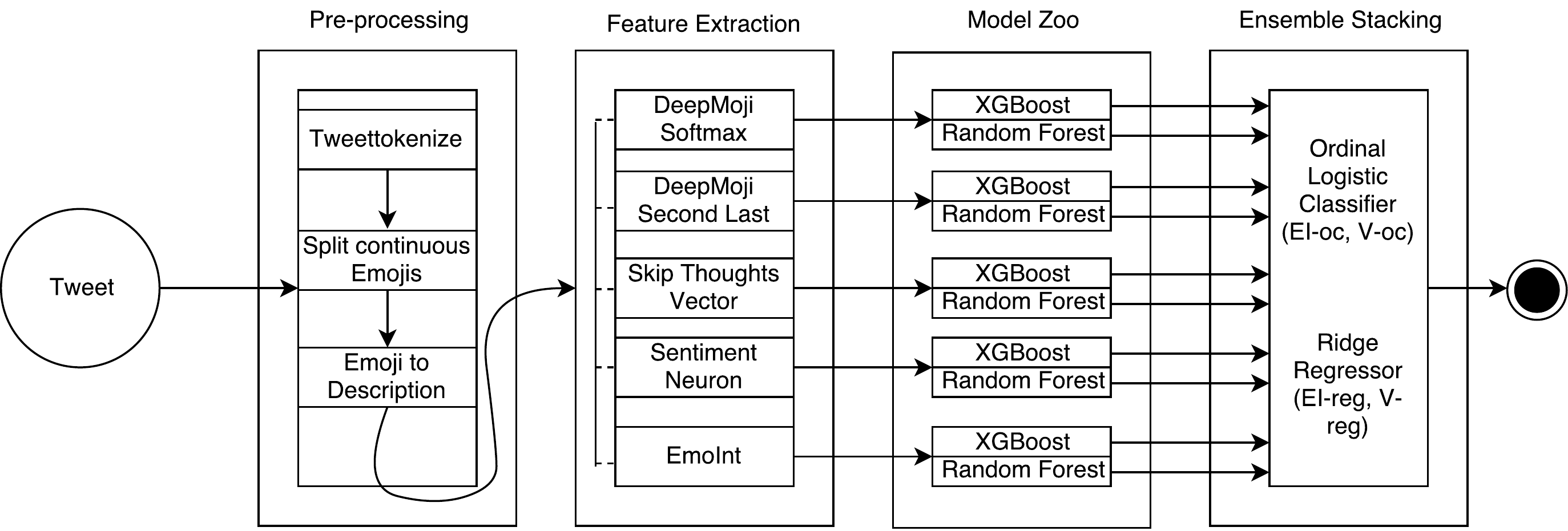}
	\caption{System Architecture.}
	\label{fig:sys_arch}
\end{figure*}

Figure \ref{fig:sys_arch} details the System Architecture. We now describe how all the different modules are tied together. The input raw tweet is pre-processed as described in Section \ref{pre-processing}. The processed tweet is passed through all the feature extractors described in Section \ref{featurizer}. At the end of this step, we extract 5 different feature vectors corresponding to each tweet. Each feature vector is passed through the model zoo where classifiers with different hyper parameters are tuned. The models are described in Section \ref{zoo}. For each vector, the results of top-2 performing models (based on cross-validation) are retained. At the end of this step, we've 10 different results corresponding to each tweet. All these results are ensembled together via stacking as described in Section \ref{stacking}. Finally, the output from the ensembler is the output returned by the system.

\subsection{Pre-processing} \label{pre-processing}
The pre-processing step modifies the raw tweets to prepare for feature extraction. Tweets are pre-processed using \textbf{tweettokenize} \footnote{https://github.com/jaredks/tweetokenize} tool. Twitter specific keywords are replaced with tokens, namely, \texttt{USERNAME}, \texttt{PHONENUMBER}, \texttt{URLs}, \texttt{timestamps}. All characters are converted to lowercase. A contiguous sequence of emojis is first split into individual emojis. We then replace an emoji with its description. The descriptions were scraped from EmojiPedia\footnote{https://emojipedia.org/}.

\subsection{Feature Extraction} \label{featurizer}
As mentioned in Section \ref{introduction}, we perform transfer learning from various state-of-the-art deep learning techniques. We will go through the following sub-sections to understand these models in detail. 

\subsubsection{DeepMoji}
DeepMoji \cite{felbo2017using} performs distant supervision on a very large dataset (1246 million tweets) comprising of noisy labels (emojis). DeepMoji was able to obtain state-of-the-art results in various downstream tasks using transfer learning. This makes it an ideal candidate for domain adaptation into related target tasks. We extract 2 different feature sets by extracting the embeddings from the softmax and the attention layer from the pre-trained DeepMoji model. The vector from softmax layer is of dimension 64 and the vector from attention layer is of dimension 2304.

\subsubsection{Skip-Thought Vectors}
Skip-Thought vectors \cite{kiros2015skip} is an off-the-shelf encoder that can produce highly generic sentence representations. Since tweets are restricted by character limit, skip-thought vectors can create a good semantic representation. This representation is then passed to the classifier. The representation is of dimension 4800.

\subsubsection{Unsupervised Sentiment Neuron}
\cite{radford2017learning} developed an unsupervised system which learned an excellent representation of sentiment. The original model was trained to generate amazon reviews, this makes the sentiment neuron an ideal candidate for transfer learning. The representation extracted from Sentiment Neuron is of size 4096.

\subsubsection{EmoInt}
Apart from all the pre-trained embeddings, we choose to also include various lexical features bundled through the EmoInt package \footnote{https://github.com/SEERNET/EmoInt} \cite{duppada2017seernet}
The lexical features include AFINN \cite{nielsen2011new}, NRC Affect Intensities \cite{mohammad2017word}, NRC-Word-Affect Emotion Lexicon \cite{mohammad2010emotions}, NRC Hashtag Sentiment  Lexicon and  Sentiment140 Lexicon \cite{mohammad2013nrc}. The final feature vector is the concatenation of all the individual features. This feature vector is of size (141, 1).

This gives us five different feature vector variants. All of these feature vectors are passed individually to the underlying models. The pipeline is explained in detail in Section \ref{pipeline}

\subsection{Machine Learning Models} \label{zoo}
We participated in 4 sub-tasks, namely, EI-oc, EI-reg, V-oc, V-reg. Two of the sub-tasks are ordinal classification and the remaining two are regressions.
We describe our approach for building ML models for both the variants in the upcoming sections.

\subsubsection{Ordinal Classification}\label{classification}
We participated in the emotion intensity ordinal classification where the task was to predict the intensity of emotions from the categories anger, fear, joy, and, sadness. Separate datasets were provided for each emotion class. 
The goal of the sub-task of valence ordinal classification was to classify the tweet into one of 7 ordinal classes [-3, 3]. We experimented with XG Boost Classifier,  Random Forest Classifier of sklearn \cite{pedregosa2011scikit}.

\subsubsection{Regression} \label{regression}
For the regression tasks (E-reg, V-reg), the goal was to predict the intensity on a scale of 0-1. We experimented with XG Boost Regressor,  Random Forest Regressor of sklearn \cite{pedregosa2011scikit}. 

The hyper-parameters of each model were tuned separately for each sub-task. The top-2 best models corresponding to each feature vector type were chosen after performing 7-fold cross-validation.

\subsubsection{Stacking} \label{stacking}
Once we get the results from all the classifiers/regressors for a given tweet, we use stacking ensemble technique to combine the results. In this case, we pass the results from the models to a meta classifier/regressor as input. The output of this meta model is treated as the final output of the system.

We observed that using ordinal regressors gave us better performance than using classifiers which treat each output class as disjoint. Ordinal Regression is a family of statistical learning methods where the output variable is discrete and ordered. We use the ordinal logistic classification with squared error \cite{rennie2005loss} from the python library Mord. \footnote{https://github.com/fabianp/mord} \cite{rennie2005loss}

In case of regression sub-tasks we observed the best cross validation results with Ridge Regression. Hence, we chose Ridge Regression as the meta regressor.

\section{Results and Analysis} \label{result}

\subsection{Task Results}

The metrics used for ranking various systems are discussed in this section.

\subsubsection{Primary Metrics}
Pearson correlation with gold labels was used as a primary metric for ranking the systems. For EI-reg and EI-oc tasks Pearson correlation is macro-averaged (MA Pearson) over the four emotion categories.

\begin{table}[t!]
\begin{center}
\begin{tabular}{|l|l|l|l|}
\hline \bf Task  &\bf Baseline & \bf 2\textsuperscript{nd} best & \bf Our Results \\ \hline
EI-reg &  0.520 & 0.776 & \textbf{0.799} \\
EI-oc  & 0.394 & 0.659 & \textbf{0.695} \\
V-reg & 0.585 & 0.861 & \textbf{0.873} \\
V-oc & 0.509 & 0.833 & \textbf{0.836} \\
\hline
\end{tabular}
\end{center}
\caption{\label{font-table} Primary metrics across various sub-tasks.}
\end{table}

Table 1 describes the results based on primary metrics for various sub-tasks in English language. Our system achieved the best performance in each of the four sub-tasks. We have also included the results of the baseline and second best performing system for comparison. As we can observe, our system vastly outperforms the baseline and is a significant improvement over the second best system, especially, in the emotion sub-tasks.

\subsubsection{Secondary Metrics}
The competition also uses some secondary metrics to provide a different perspective on the results. Pearson correlation for a subset of the test set that includes only those tweets with intensity score greater or equal to 0.5 is used as the secondary metric for the regression tasks. For ordinal classification tasks following secondary metrics were used:

\begin{itemize}

\item Pearson correlation for a subset of the test set that includes only those tweets with intensity classes low X, moderate X, or high X (where X is an emotion). The organizers refer to this set of tweets as the some-emotion subset (SE). 
\item Weighted quadratic kappa on the full test set
\item Weighted quadratic kappa on the some-emotion subset of the test set

\end{itemize}

The results for secondary metrics are listed in Table 2 and 3. We have also included the ranking in brackets along with the score. We see that our system achieves the top rank according to all the secondary metrics, thus, proving its robustness. 

\begin{table}[t!]
\begin{center}
\begin{tabular}{|l|l|l|l|}
\hline \bf Task  &\bf 	Pearson (SE) & \bf Kappa & \bf Kappa (SE)  \\ \hline
V-oc & 0.884 (1) & 0.831 (1) & 0.873 (1)\\
EI-oc & 0.547 (1) & 0.669 (1) & 0.503 (1)\\
\hline
\end{tabular}
\end{center}
\caption{\label{font-table} Secondary metrics for ordinal classification sub-tasks. System rank is mentioned in the brackets.}
\end{table}

\begin{table}[t!]
\begin{center}
\begin{tabular}{|l|l|}
\hline \bf Task  &\bf 	Pearson (gold in 0.5-1) \\ \hline
V-reg & 0.697 (1) \\
EI-reg & 0.638 (1) \\
\hline
\end{tabular}
\end{center}
\caption{\label{font-table} Secondary metrics for regression sub-tasks.  System rank is mentioned in brackets.}
\end{table}

\subsection{Feature Importance}
The performance of the system is highly dependent on the discriminative ability of the tweet representation generated by the featurizers. We measure the predictive power for each of the featurizer used by calculating the pearson correlation of the system using only that featurizer. We describe the results for each sub task separately in tables 4-7.

\begin{table}[t!]
\begin{center}
\begin{tabular}{|l|l|}
\hline \bf Feature Set  &\bf Pearson \\ \hline
Deepmoji (softmax layer)& 0.808 \\
Deepmoji (attention layer)& 0.843 \\
EmoInt & 0.823 \\
Unsupervised sentiment Neuron & 0.714 \\
Skip-Thought Vectors & 0.777 \\
\hline
Combined & \textbf{0.873} \\
\hline
\end{tabular}
\end{center}
\caption{\label{font-table} Pearson Correlation for V-reg task. Best results are highlighted in bold. }
\end{table}

\begin{table}[t!]
\begin{center}
\begin{tabular}{|l|l|}
\hline \bf Feature Set  &\bf Pearson \\ \hline
Deepmoji (softmax layer)& 0.780 \\
Deepmoji (attention layer)& 0.813 \\
EmoInt & 0.785 \\
Unsupervised sentiment Neuron & 0.685 \\
Skip-Thought Vectors & 0.748 \\
\hline
Combined & \textbf{0.836} \\
\hline
\end{tabular}
\end{center}
\caption{\label{font-table} Pearson Correlation for V-oc task. Best results are highlighted in bold. }
\end{table}

\begin{table}[t!]
\begin{center}
\begin{tabular}{|l|l|}
\hline \bf Feature Set  &\bf Pearson \\ \hline
Deepmoji (softmax layer)& 0.703  \\
Deepmoji (attention layer)& 0.756  \\
EmoInt & 0.694  \\
Unsupervised sentiment Neuron & 0.548 \\
Skip-Thought Vectors & 0.656 \\
\hline
Combined & \textbf{0.799}\\
\hline
\end{tabular}
\end{center}
\caption{\label{font-table}  Macro-Averaged Pearson Correlation for EI-reg task. Best results are highlighted in bold.}
\end{table}

\begin{table}[t!]
\begin{center}
\begin{tabular}{|l|l|}
\hline \bf Feature Set  &\bf Pearson\\ \hline
Deepmoji softmax layer & 0.611 \\
Deepmoji attention layer& 0.664 \\
EmoInt & 0.596 \\
Unsupervised sentiment Neuron & 0.445 \\
Skip-Thought Vectors & 0.557 \\
\hline
Combined & \textbf{0.695} \\
\hline
\end{tabular}
\end{center}
\caption{\label{font-table} Macro-Averaged Pearson Correlation for EI-oc task. Best results are highlighted in bold. }
\end{table}

We observe that deepmoji featurizer is the most powerful featurizer of all the ones that we've used. Also, we can see that stacking ensembles of models trained on the outputs of multiple featurizers gives a significant improvement in performance.

\subsection{System Limitations}
We analyze the data points where our model's prediction is far from the ground truth. We observed some limitations of the system, such as, sometimes understanding a tweet's requires contextual knowledge about the world. Such examples can be very confusing for the model. We use deepmoji pre-trained model which uses emojis as proxy for labels, however partly due to the nature of twitter conversations same emojis can be used for multiple emotions, for example, joy emojis can be sometimes used to express joy, sometimes for sarcasm or for insulting someone. One such example is \textit{'Your club is a laughing stock'}. Such cases are sometimes incorrectly predicted by our system.

\section{Future Work \& Conclusion} \label{conclusion}
The paper studies the effectiveness of various representations of tweets and proposes ways to combine them to obtain state-of-the-art results. We also show that stacking ensemble  of various classifiers learnt using different representations can vastly improve the robustness of the system.

Further improvements can be made in the pre-processing stage. Instead of discarding various tokens such as punctuation's, incorrectly spelled words, etc, we can utilize the information by learning their semantic representations. Also, we can improve the system performance by employing multi-task learning techniques as various emotions are not independent of each other and information about one emotion can aid in predicting the other. Furthermore, more robust techniques can be employed for distant supervision which are less prone to noisy labels to get better quality training data.

\bibliography{semeval2018}
\bibliographystyle{acl_natbib}
\end{document}